\documentclass[letterpaper, 10 pt, conference]{ieeeconf}
\IEEEoverridecommandlockouts
\usepackage{amsmath,amssymb,amsfonts}
\usepackage{graphicx}
\usepackage{xcolor}
\usepackage[hyphens]{url}
\usepackage[hidelinks,pdftex]{hyperref}
\usepackage[all]{hypcap}
\usepackage[T1]{fontenc}
\usepackage{multirow}
\usepackage{algorithmic}
\usepackage{cite}
\usepackage{textcomp}
\usepackage{subcaption}
\usepackage{balance}

\begin{document}
\hypersetup{
	pdfinfo ={
			Title={Sim2Dust: Mastering Dynamic Waypoint Tracking on Granular Media},
			Subject={2025 International Conference on Space Robotics (iSpaRo)},
			Author={Andrej Orsula, Matthieu Geist, Miguel Olivares-Mendez, Carol Martinez},
			Creator={LaTeX},
			Keywords={Intelligent and autonomous space robotics systems; Planetary exploration; Space robotic locomotion and terramechanics},
		}
}

\title{\textbf{Sim2Dust: Mastering Dynamic Waypoint Tracking on Granular Media}}


\author{
	Andrej Orsula\textsuperscript{1}
	\and
	Matthieu Geist\textsuperscript{2}
	\and
	Miguel Olivares-Mendez\textsuperscript{1}
	\and
	Carol Martinez\textsuperscript{1}
}

\maketitle\thispagestyle{empty}\pagestyle{empty}

\footnotetext[1]{
	Space Robotics Research Group~(SpaceR),
	Interdisciplinary Centre for Security, Reliability and Trust~(SnT),
	University of Luxembourg
		{\tt andrej.orsula@uni.lu}
}
\footnotetext[2]{
	Earth Species Project
}

\begin{abstract}
    Reliable autonomous navigation across the unstructured terrains of distant planetary surfaces is a critical enabler for future space exploration. However, the deployment of learning-based controllers is hindered by the inherent sim-to-real gap, particularly for the complex dynamics of wheel interactions with granular media. This work presents a complete sim-to-real framework for developing and validating robust control policies for dynamic waypoint tracking on such challenging surfaces. We leverage massively parallel simulation to train reinforcement learning agents across a vast distribution of procedurally generated environments with randomized physics. These policies are then transferred zero-shot to a physical wheeled rover operating in a lunar-analogue facility. Our experiments systematically compare multiple reinforcement learning algorithms and action smoothing filters to identify the most effective combinations for real-world deployment. Crucially, we provide strong empirical evidence that agents trained with procedural diversity achieve superior zero-shot performance compared to those trained on static scenarios. We also analyze the trade-offs of fine-tuning with high-fidelity particle physics, which offers minor gains in low-speed precision at a significant computational cost. Together, these contributions establish a validated workflow for creating reliable learning-based navigation systems, marking a substantial step towards deploying autonomous robots in the final frontier.
\textit{The source code is available at~\href{https://github.com/AndrejOrsula/space_robotics_bench}{https://github.com/AndrejOrsula/space\_robotics\_bench}.}

\end{abstract}

\section{Introduction}

The next era of space exploration aims to establish a sustained human presence beyond Earth through ambitious programs like Artemis~\cite{nasa2020artemis}. This vision requires a new generation of highly autonomous robotic systems to perform essential tasks in remote and hazardous locations. Within these fleets, wheeled rovers will serve as key enablers for these future off-world outposts. They will traverse uncharted landscapes to conduct geological surveys, transport resources, and aid in constructing infrastructure~\cite{jpl2024enabling,zhang2019progress}. The success of these long-duration missions is therefore directly linked to the ability of rovers to navigate reliably with minimal human supervision.

A fundamental obstacle to this autonomy is the nature of extraterrestrial terrain. Planetary surfaces are profoundly unstructured and often covered by a layer of fine-grained granular material called regolith. The interaction dynamics between a rover's wheels and this deformable medium are difficult to model. Phenomena such as wheel slippage and sinkage introduce significant uncertainty that limits the efficacy of traditional control methods. Data-driven approaches like reinforcement learning~(RL) offer a compelling alternative. An RL agent can learn a complex control policy through trial and error, while implicitly capturing the physics of its environment without an explicit analytical model~\cite{sutton2018reinforcement}.

However, the primary barrier to applying RL in this domain is the infeasibility of training directly on a celestial body. The risk, cost, and limited bandwidth associated with space missions make real-world data collection impractical. Simulation is therefore the only viable training ground. This reliance introduces the critical sim-to-real gap. A policy trained in a virtual environment often fails when transferred to the physical world due to subtle discrepancies between the systems. This gap is especially pronounced for tasks dominated by complex contact dynamics, such as traversal on granular surfaces.

\begin{figure}[t]
    \vspace{2.057mm}
    \centering
    \includegraphics[width=\linewidth]{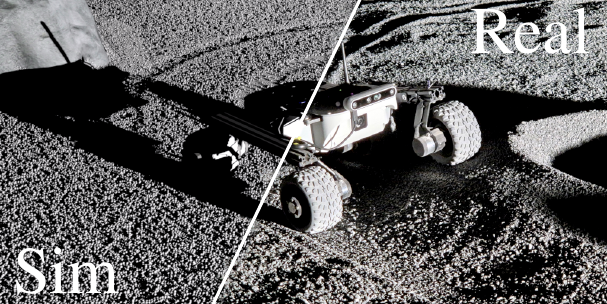}
    \caption{Agents are trained to track dynamic waypoints in procedurally generated scenarios of the Space Robotics Bench. The generalization learned from diverse experience enables the acquired policies to be transferred zero-shot to a physical rover on granular media in a lunar-analogue facility.}
    \label{fig:concept}
    \vspace{-1.0em}
\end{figure}

This work confronts the sim-to-real challenge for rover navigation. We argue that creating a singular perfect digital twin is often impractical or even impossible due to the limited knowledge of extraterrestrial environments. Instead, our approach leverages procedural generation to fabricate a vast distribution of diverse simulated scenarios. Training agents across this wide range of conditions forces them to learn generalizable strategies that are robust to real-world variations. This learned robustness enables successful sim-to-real transfer to physical rovers, as illustrated in Fig.~\ref{fig:concept}.

We present a complete sim-to-real framework for dynamic waypoint tracking on granular media using the Space Robotics Bench~\cite{orsula2025space}. We systematically compare several RL algorithms and action smoothing filters to find the most effective strategies for hardware deployment. We provide strong empirical evidence that training with procedural diversity is critical for successful zero-shot transfer. We further show that fine-tuning with high-fidelity particle physics can harden a policy against unmodeled terramechanic effects. Together, these contributions establish a validated workflow for developing reliable learning-based autonomy in the context of space robotics.

\section{Related Work}

This research integrates concepts from three distinct domains. These are the study of vehicle motion on granular media, the application of robot learning to navigation, and the use of simulation for the development of space systems.

\subsection{Traversal on Granular Media}

Terramechanics, the scientific study of vehicle mobility on deformable terrain, provides the physical basis for our work. Foundational studies have established analytical models to predict wheel slippage and steering forces~\cite{ishigami2007terramechanics}. These models later informed path planners that could evaluate paths based on vehicle stability and energy use~\cite{ishigami2007path,ishigami2011path}. While insightful, these classical models require extensive empirical data and struggle with the high variability of natural terrain. Modern physics engines like Chrono~\cite{tasora2015chrono} enable high-fidelity simulation of these interactions and have been used to train rovers for difficult navigation tasks~\cite{xu2024reinforcement}. Such approaches advance physical modeling but depend on an accurate representation of the environment. Our work takes a complementary path. We use learning to implicitly capture complex wheel-terrain dynamics, aiming for a policy that is robust to the inaccuracies of any single model.

\subsection{Learning-Based Robot Navigation}

RL has emerged as a powerful tool for solving complex robot navigation problems. For planetary rovers, learning has been applied to diverse tasks like collaborative path planning~\cite{lu2024lunar}, navigation on extreme slopes~\cite{xu2024reinforcement}, and developing fault-tolerant controllers~\cite{park2023deep}. A central theme in this research is achieving robust generalization. Approaches for this challenge include making policies resilient to sensor noise through multi-stage training~\cite{mortensen2024twostage} and training agents in procedurally generated environments~\cite{cobbe2020leveraging}. The latter has shown particular promise, though its exploration for rovers has been limited to simplified 2D simulations~\cite{koutras2021marsexplorer}. While these studies advance autonomous navigation, most focus on traversal over rigid surfaces. Our work contributes to this field by specifically tackling policy learning for dynamic waypoint tracking directly on granular media. The interaction physics of this domain are particularly challenging to simulate, representing a critical step for creating truly robust navigation systems.

\subsection{Simulation-Centric Learning for Space Robotics}

The extreme cost and logistical challenges of space missions make simulation an indispensable tool for verifying spacecraft dynamics~\cite{hughes2014verification,kenneally2020basilisk}. The rise of RL has led to a new generation of simulators for specific space applications, including rover navigation~\cite{mortensen2024rlroverlab}, spacecraft rendezvous~\cite{el2024drift}, and in-orbit manipulation~\cite{wang2022collision}. A key technique for policy transfer is domain randomization, where simulation properties are varied during training~\cite{tobin2017domain}. This method forces the policy to become more invariant to the sim-to-real gap, which in turn improves the transfer success~\cite{bousmalis2018using}. Our research builds on these principles by using the Space Robotics Bench~(SRB) as our development platform~\cite{orsula2025space}. While SRB is a comprehensive framework for space robotics research and supports several robot morphologies, this paper presents its first application to a complete sim-to-real rover navigation workflow. We use its capabilities for parallel simulation and procedural diversity to systematically investigate the factors that enable successful zero-shot transfer on granular media.

\section{Framework for Sim-to-Real Rover Autonomy}

Our methodology for creating and deploying robust rover autonomy rests on an integrated framework. This system combines a powerful simulation environment for training, a realistic physical testbed for validation, and a mission-relevant control task.

\subsection{Simulation Framework: Space Robotics Bench}

All policy development occurs within SRB~\cite{orsula2025space}, our open-source framework built on NVIDIA Isaac Lab~\cite{mittal2023orbit}. This backend provides GPU-accelerated physics and rendering, which enables the massive parallelization required for modern RL workflows to collect diverse experience efficiently. Crucially for this work, the framework integrates procedural content generation~(PCG) pipelines for creating varied terrains and supports extensive domain randomization of physical parameters to bridge the sim-to-real gap.

\begin{figure}[b]
    \vspace{-0.35em}
    \centering
    \includegraphics[width=\linewidth]{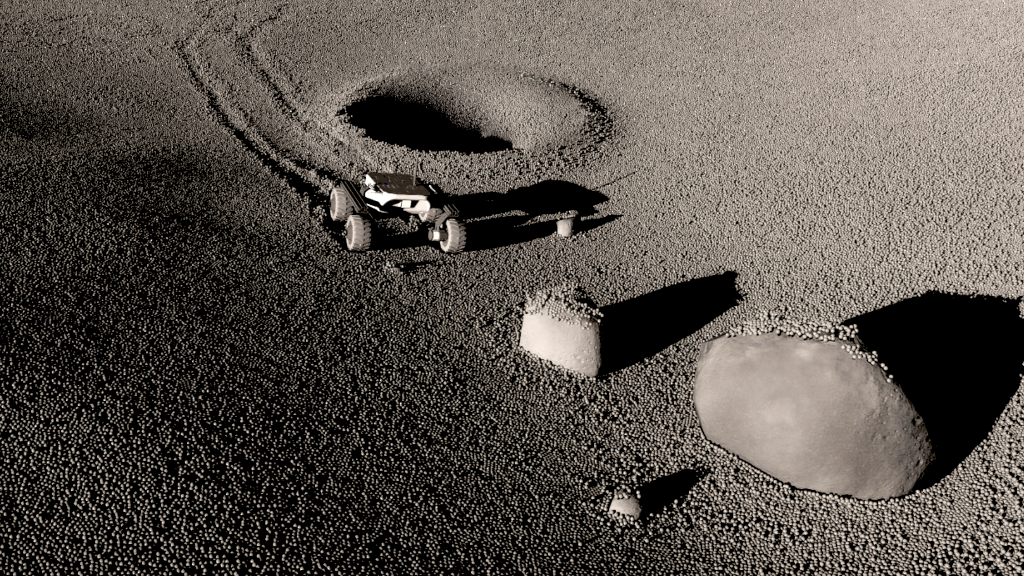}
    \caption{Visualization of our high-fidelity simulation environment. A procedurally generated terrain mesh is populated with millions of discrete particles that enable a more realistic simulation of complex wheel-regolith interaction dynamics.}
    \label{fig:sim_with_particles}
    \vspace{-0.35em}
\end{figure}

\begin{figure}[t]
    \vspace{0.525em}
    \centering
    \subcaptionbox{Stacked regime.\label{fig:sim_stacked}}{%
        \includegraphics[width=0.4935\linewidth]{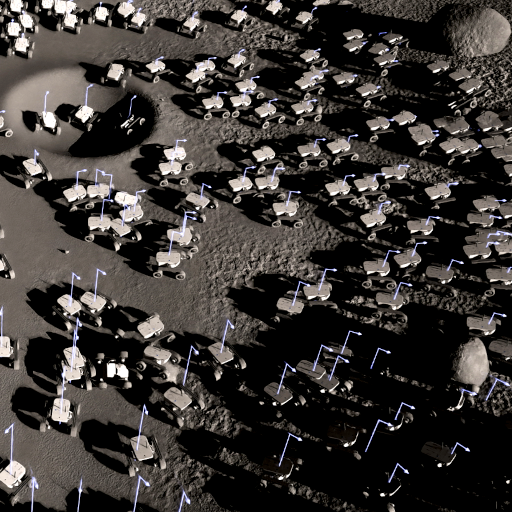}}
    \hfill
    \subcaptionbox{Procedural regime.\label{fig:sim_diverse}}{%
        \includegraphics[width=0.4935\linewidth]{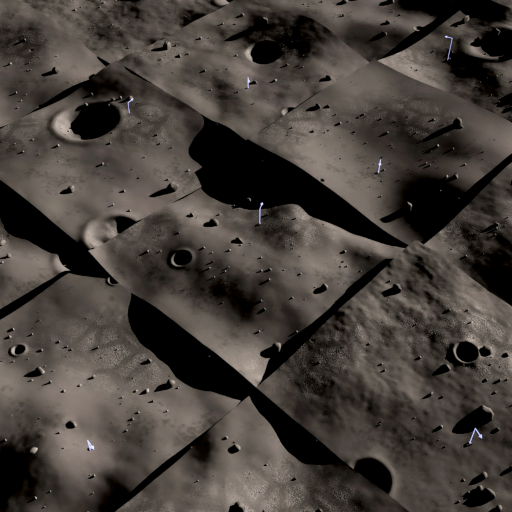}}
    \caption{SRB supports massively parallel simulation in two primary regimes. In the stacked regime, all environment instances are superimposed and share a single static terrain, which risks policy overfitting. In contrast, the procedural regime exposes each instance to a unique procedurally generated terrain to foster robustness and generalization. Blue arrows indicate the dynamically evolving target waypoints.}
    \label{fig:sim_regimes}
    \vspace{-1.0em}
\end{figure}

The framework supports high-fidelity physics, including the simulation of millions of discrete particles to model granular media, as shown in Fig.~\ref{fig:sim_with_particles}. However, such detailed simulations are computationally expensive. This cost limits their use for large-scale policy training. Therefore, our primary training methodology leverages the massive parallelization of SRB on rigid lunar surfaces. We run hundreds of environments simultaneously to generate the vast amount of experience required for RL. To ensure this experience is sufficiently diverse, we rely on procedural generation to create a near-infinite variety of terrains. As illustrated in Fig.~\ref{fig:sim_regimes}, we explore two distinct training regimes. In the stacked training regime (Fig.~\ref{fig:sim_stacked}), all agents are trained on a single procedurally generated environment that is shared among them. This setup risks policy overfitting to the specific features of that one terrain. In contrast, the diverse procedural regime (Fig.~\ref{fig:sim_diverse}) provides each agent with its own unique procedurally generated terrain, which is central to our hypothesis that exposing agents to a wide distribution of environments is crucial for learning generalization.

Our procedural terrains follow a sequential process to create diverse landscapes. The PCG pipeline begins by displacing a base mesh with low-frequency Perlin noise to form the primary topography. Subsequent layers of higher-frequency noise are applied to create more detailed features. For example, Voronoi noise forms the rims of impact craters. The parameters of each layer are then randomized to generate a wide distribution of scenes. Lastly, procedural boulders and rocks are scattered across the terrain using Poisson disk sampling to ensure a natural and plausible distribution.

Beyond PCG, we employ extensive domain randomization to enhance policy robustness. At the start of each training episode, we vary key simulation parameters to prepare the policy for real-world unpredictability. These include environmental properties such as the gravity vector and platform-specific adjustments like small offsets of the rover's base frame to account for manufacturing variations and calibration errors. Furthermore, we inject randomized noise and variable delays into both actions and observations. This process builds resilience to the unpredictable latencies and sensor inaccuracies inherent in a physical system. To complete the framework, SRB packages the final trained policy into a standard ROS~2 interface~\cite{macenski2022ros2}, providing a seamless pathway for zero-shot deployment on the physical hardware.

\subsection{Lunar Testbed: LunaLab}

\begin{figure}[b]
    \vspace{-0.45em}
    \centering
    \includegraphics[width=\linewidth]{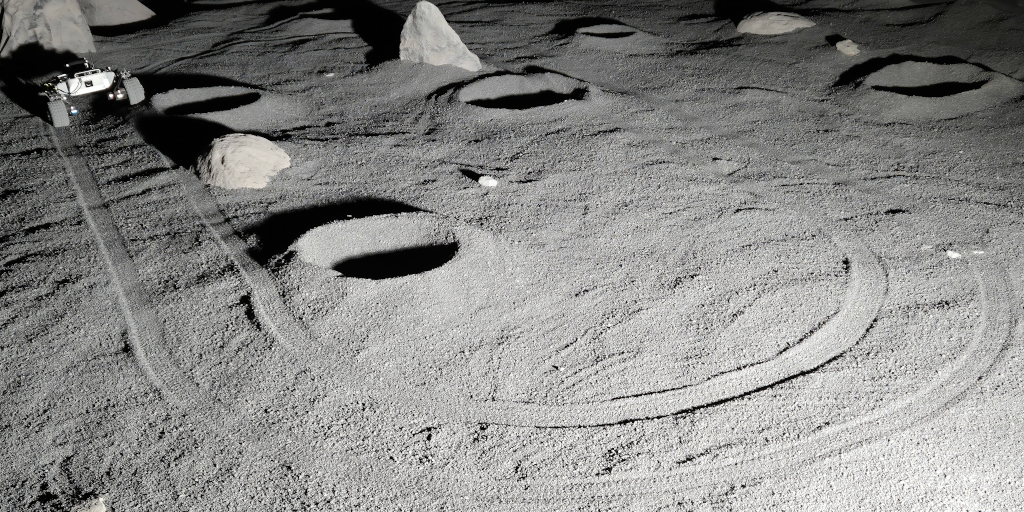}
    \caption{Real-world validation is performed with a Leo Rover inside the LunaLab facility~\cite{ludivig2020building}, which serves as a lunar-analogue testbed filled with basalt gravel. It is equipped with a Sun emulator and a motion capture system for ground-truth state estimation during policy execution and evaluation.}
    \label{fig:lunalab_capsule}
    \vspace{-0.35em}
\end{figure}

We conduct all real-world validation in LunaLab, a lunar-analogue facility at the University of Luxembourg containing 20 tons of basalt gravel that represents the properties of regolith~\cite{ludivig2020building}. Our robotic platform is Leo Rover, a four-wheeled skid-steer mobile robot with a wheelbase of 29.5~cm, as depicted in Fig.~\ref{fig:lunalab_capsule}. For ground-truth localization, we use an OptiTrack motion capture system that provides high-frequency pose data. This data serves a dual role, as it supplies the real-time state for interpreting global waypoints in the robot's local frame and logs the complete trajectory for analysis. This setup is critical because it allows us to evaluate the policy's performance independent of any potential state estimation errors.

\subsection{Task: Dynamic Waypoint Tracking}

The objective of the agent is to master dynamic waypoint tracking. We formulate this as a partially observable Markov decision process compatible with the Gymnasium API~\cite{towers2024gymnasium}. The goal is to control the rover to continuously and accurately follow a moving target pose within the environment.

The policy operates at a 25~Hz control frequency. At each timestep, the agent receives a command vector containing the relative 2D position to the target waypoint and a 2D vector representing the sine and cosine of the relative yaw orientation error. To better simulate real-world conditions, we inject two forms of noise into the observations. A per-episode noise, sampled from $\mathcal{N}(0.0, [1.0\ \text{cm}, 2.5^{\circ}])$, models persistent biases from mechanical misalignments or calibration errors. A per-timestep noise, sampled from $\mathcal{N}(0.0, [0.25\ \text{cm}, 0.5^{\circ}])$, represents transient signal jitter.

The action space is a 2D continuous vector for the desired linear and angular velocity commands. These actions are normalized and mapped to the rover's maximum linear speed of 40~cm/s and angular speed of 60\textdegree/s. To account for hardware and communication latencies, we introduce randomized delays during training. We delay observations by up to 40~ms (1~step) and actions by up to 120~ms (3~steps). The specific delay for each environment instance is randomized per episode and has a 1\% chance of changing every second to model the unpredictable latencies.

The reward function is carefully shaped to encourage a specific sequence of behaviors for effective tracking. A continuous penalty on the Euclidean distance and a reward for pointing towards the target guide the general approach. As the rover nears the target, additional rewards for precise position and orientation alignment become dominant. Once the rover is correctly aligned at the waypoint, a final reward term encourages it to minimize its action rate in order to promote a stable stop or smooth tracking behavior. A small, persistent penalty on large action changes is also included to discourage jerky movements throughout the episode. To ensure the policy generalizes, the target waypoint itself follows a unique, smoothly randomized trajectory in each parallel environment instance with a 60~s truncation window.

\section{Experimental Results}

We conduct a series of systematic experiments to investigate the factors that enable successful sim-to-real transfer for rover navigation. We specifically evaluate the impact of the learning algorithm, training diversity, high-fidelity physics, and action smoothing on policy performance and robustness.

\subsection{Experimental Protocol}

Our core evaluation methodology is zero-shot transfer. Policies are trained exclusively in simulation and deployed directly to the physical Leo Rover. For each experimental configuration, we train five independent seeds. To ensure a representative comparison and mitigate the effects of outlier training runs, we select the agent with the median final episodic return (last 1M steps) for real-world deployment.

We assess real-world performance by commanding the rover to follow a pre-configured looping \textit{capsule} trajectory. All reported data is recorded after an agent completes its first lap to reduce the impact of initial conditions. Furthermore, we evaluate policy robustness across a range of dynamic conditions by setting the target waypoint to move at three constant velocities:~5,~15,~and 25~cm/s.

Average tracking error~(ATE) is our primary metric to quantify performance and the ability of agents to match their target velocity. A lower ATE signifies a more precise and stable policy. We compute the position error as the mean Euclidean distance between the rover and the dynamic target. The orientation error is the mean absolute difference between the rover's heading and the target's heading. Additional experiment-specific metrics are also reported.

\subsection{Algorithmic Comparison}

\begin{figure}[t]
    \vspace{0.525em}
    \centering
    \includegraphics[width=\linewidth]{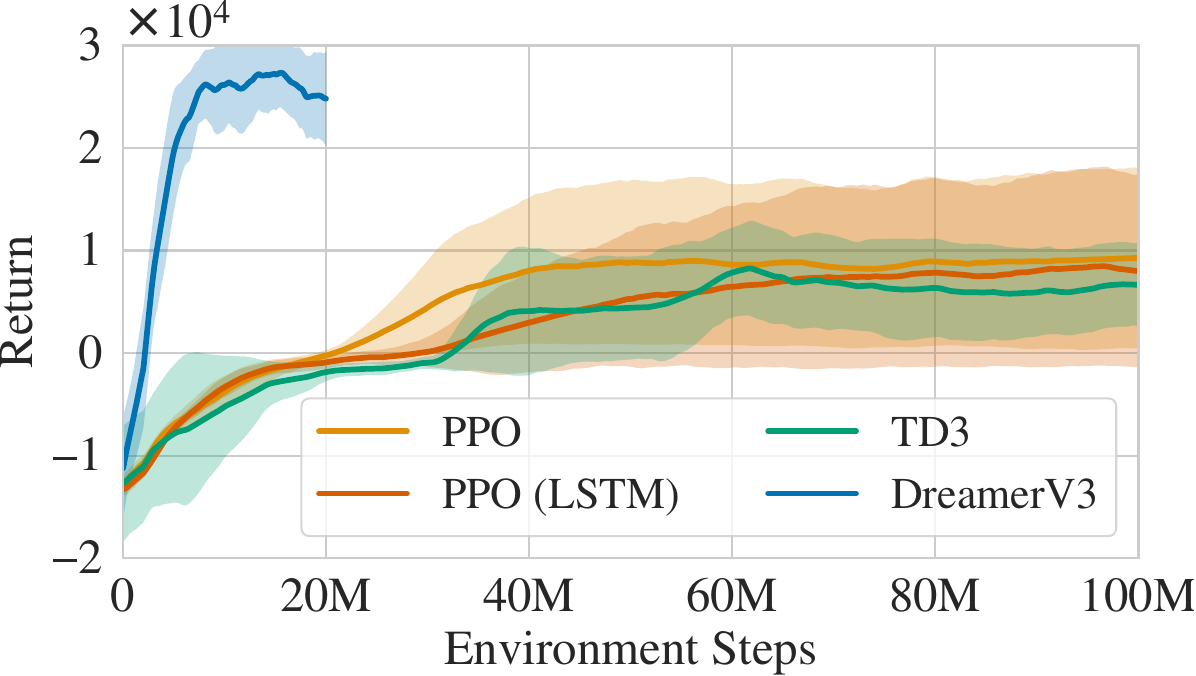}
    \caption{Learning curves of RL algorithms during the training in SRB simulation, averaged over five random seeds, with shaded regions representing the standard deviation.}
    \label{fig:learning_curves}
    \vspace{-1.0em}
\end{figure}

Our first experiment identifies the most suitable RL algorithm for the dynamic waypoint tracking task. We evaluate three algorithms that represent distinct learning paradigms. Proximal Policy Optimization~(PPO)~\cite{schulman2017proximal} is a standard on-policy algorithm, and we also test a variant with a Long Short-Term Memory~(LSTM)~\cite{hochreiter1997long}. Twin Delayed Deep Deterministic Policy Gradient~(TD3)~\cite{fujimoto2018addressing} represents modern off-policy methods. Finally, DreamerV3~\cite{hafner2025mastering} is selected as a representative model-based algorithm due to its demonstrated performance across diverse tasks. Each agent was trained in 512 unique parallel environment instances as depicted in Fig.~\ref{fig:sim_diverse}. The model-free agents were trained for 100M steps, while the sample-efficient DreamerV3 was trained for 20M steps. The learning curves in Fig.~\ref{fig:learning_curves} show that DreamerV3 achieves a higher final episodic return in significantly fewer steps, which indicates its superior performance and sample efficiency. Hyperparameters are listed in the Appendix.

\begin{figure}[hb]
    \centering
    \vspace{-1.15em}
    \includegraphics[width=0.98\linewidth]{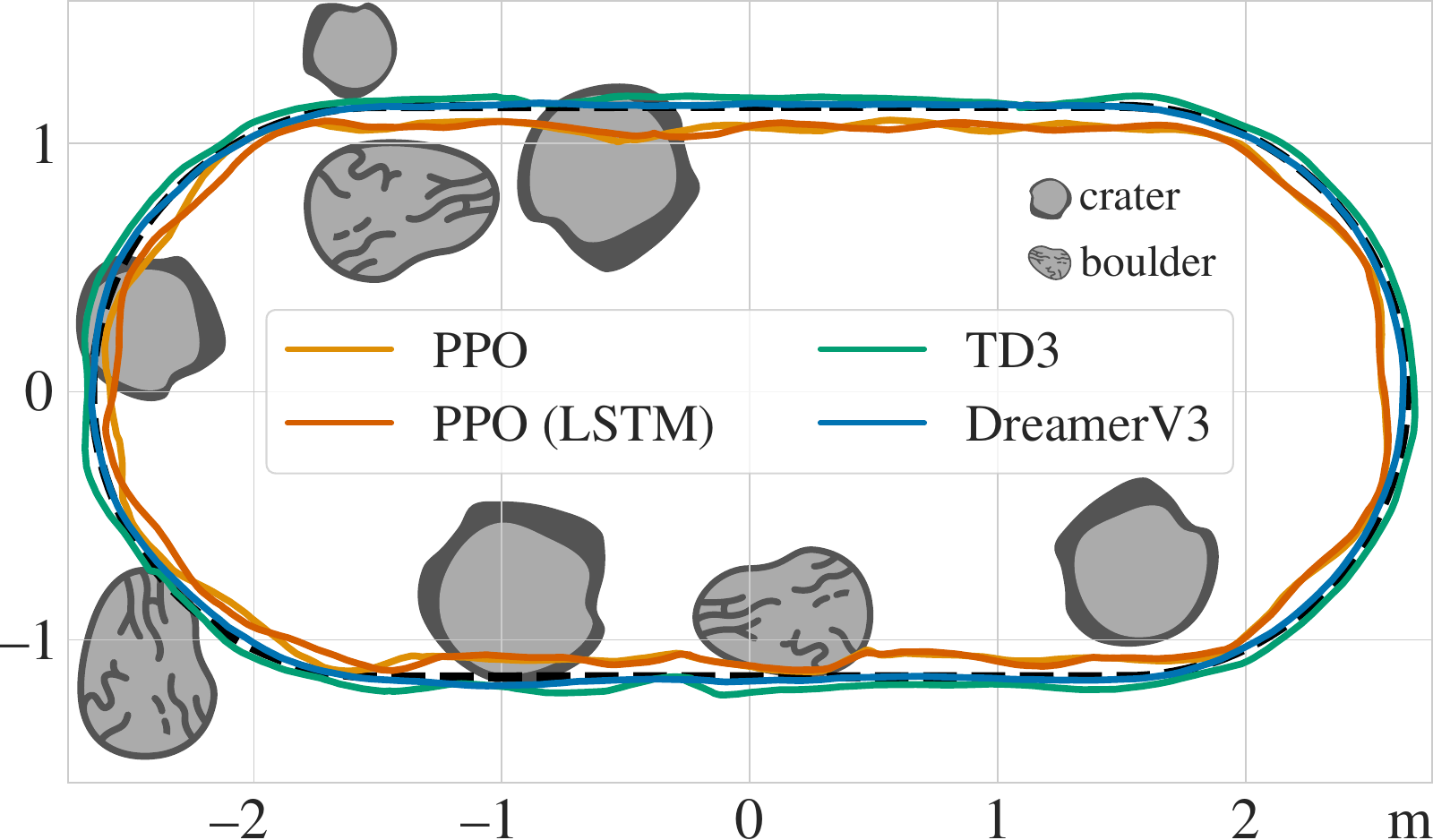}
    \caption{Real-world \textit{capsule} trajectory for policies trained with different RL algorithms. Major environmental obstacles (craters and boulders) are graphically indicated for context.}
    \label{fig:traj_algos}
    \vspace{-0.35em}
\end{figure}

The quantitative sim-to-real transfer results are presented in \textsc{Table~\ref{tab:results_algos}}. The DreamerV3 agent achieves a substantially lower ATE across all tested velocities. This superior accuracy is qualitatively confirmed in Fig.~\ref{fig:traj_algos}, which illustrates the real-world trajectories of the different agents. The path traced by the DreamerV3 policy is visibly smoother and more closely aligned with the target trajectory, while other agents exhibit larger deviations and more erratic behavior. The table also highlights the computational trade-offs. While PPO has a significantly lower inference latency, the sample efficiency of DreamerV3 allows it to complete its training in a comparable amount of time on a single NVIDIA RTX 4090 GPU.

\begin{table}[hb]
    \vspace{0.5em}
    \centering
    \caption{\textsc{Sim-to-Real Performance, Training Duration, and Inference Latency of RL Algorithms}}
    \label{tab:results_algos}
    \resizebox{\linewidth}{!}%
    {%
        \addtolength{\tabcolsep}{-0.3em}
        \begin{tabular}{@{}r|cccc@{}}
                                      & \textbf{PPO}                                                    & \textbf{PPO (LSTM)}                           & \textbf{TD3}                                  & \textbf{DreamerV3}                            \\
            \hline
            5  cm/s                   & 13.2 cm \vline\ 7.8\textdegree                                  & 11.4 cm \vline\ 4.8\textdegree                & 12.6 cm \vline\ 8.5\textdegree                & \textbf{2.3 cm \vline\ 1.7\textdegree}        \\
            15 cm/s                   & 13.7 cm \vline\ 8.6\textdegree                                  & 11.2 cm \vline\ 8.1\textdegree                & 11.6 cm \vline\ 6.2\textdegree                & \textbf{3.3 cm \vline\ 1.9\textdegree}        \\
            25 cm/s                   & 14.8 cm \vline\ 8.7\textdegree                                  & 12.9 cm \vline\ 9.9\textdegree                & 13.1 cm \vline\ 9.1\textdegree                & \textbf{3.6 cm \vline\ 2.3\textdegree}        \\
            \hline
            Training                  & 13h30 (100M)                                                    & 25h00 (100M)                                  & 15h00 (100M)                                  & 17h30 (20M)                                   \\
            \hline
            Infer|GPU                 & $\mathbf{0.42\hspace{-0.5mm}\pm\hspace{-0.5mm}0.1}$ \textbf{ms} & $0.71\hspace{-0.5mm}\pm\hspace{-0.5mm}0.2$ ms & $0.43\hspace{-0.5mm}\pm\hspace{-0.5mm}0.1$ ms & $1.27\hspace{-0.5mm}\pm\hspace{-0.5mm}0.1$ ms \\
            Infer|\hspace{0.155mm}CPU & $\mathbf{0.24\hspace{-0.5mm}\pm\hspace{-0.5mm}0.1}$ \textbf{ms} & $0.71\hspace{-0.5mm}\pm\hspace{-0.5mm}0.2$ ms & $0.43\hspace{-0.5mm}\pm\hspace{-0.5mm}0.1$ ms & $2.38\hspace{-0.5mm}\pm\hspace{-0.5mm}0.2$ ms \\
        \end{tabular}
    }
    \vspace{-1.15em}
\end{table}

\begin{figure*}[t]
    \vspace{0.525em}
    \centering
    \includegraphics[width=\linewidth]{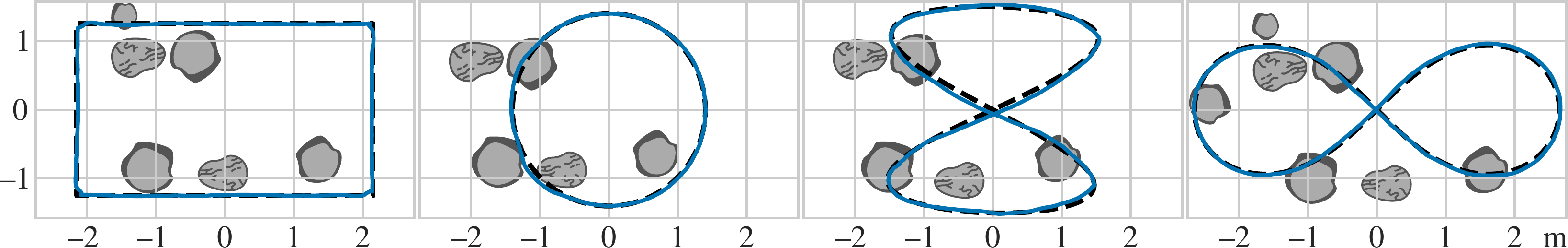}
    \caption{Additional real-world trajectories (\textit{rectangle}, \textit{circle}, \textit{Lissajous}, and \textit{lemniscate}) executed by the DreamerV3 agent.}
    \label{fig:traj_showcase}
    \vspace{-1.0em}
\end{figure*}

\begin{figure}[t]
    \vspace{0.425em}
    \centering
    \includegraphics[width=\linewidth]{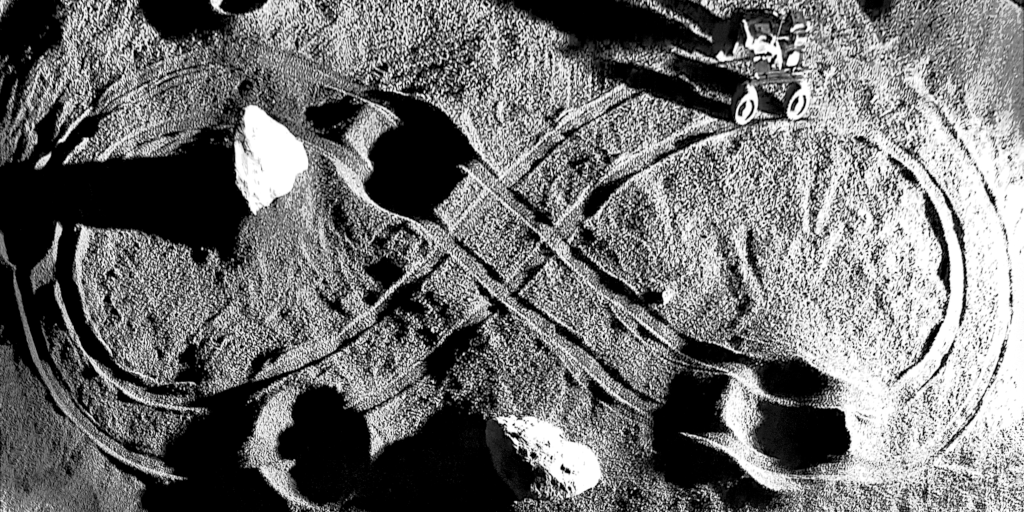}
    \caption{The \textit{lemniscate} path imprinted by the rover's wheels during a real-world deployment with the DreamerV3 agent.}
    \label{fig:lunalab_lemniscate}
    \vspace{-1.0em}
\end{figure}

We further showcase the robustness and generalization of the DreamerV3 agent by deploying it on a series of more complex, unseen paths at 15~cm/s. As shown in Fig.~\ref{fig:traj_showcase}, the agent successfully follows diverse trajectories using the same zero-shot transfer workflow. The precision of the learned controller is further underscored by the physical tracks it leaves in the granular medium. Figure~\ref{fig:lunalab_lemniscate} shows the clean and repeatable \textit{lemniscate} path imprinted by the rover's wheels at 25~cm/s, which demonstrates the stability and reliability of the policy. Given its clear advantages in both performance and sample efficiency, we use DreamerV3 for the remainder of our experiments.

\subsection{Generalization through Simulation Diversity and Fidelity}

This experiment systematically evaluates our core hypothesis that policy robustness is a direct result of the diversity and fidelity of the simulation environment. We analyze the four distinct training regimes presented in \textsc{Table~\ref{tab:results_regime}} to determine the most effective strategy for zero-shot sim-to-real transfer.

\begin{table}[b]
    \vspace{-0.5em}
    \centering
    \caption{\textsc{Sim-to-Real Performance and Training Duration of Different Regimes}}
    \label{tab:results_regime}
    \resizebox{\linewidth}{!}%
    {%
        \addtolength{\tabcolsep}{-0.15em}
        \begin{tabular}{@{}r|cccc@{}}
                     & \textbf{Static}               & \textbf{DR}                   & \textbf{DR\&PCG}                                       & \textbf{DR\&PCG+PF}                                     \\
            \hline
            5  cm/s  & 3.4 cm \vline\ 4.2\textdegree & 2.5 cm \vline\ 1.6\textdegree & 2.3 cm \hspace{0.225mm}\textbf{\vline}\ 1.7\textdegree & \textbf{2.2 cm \hspace{-0.225mm}\vline\ 1.5\textdegree} \\
            15 cm/s  & 4.2 cm \vline\ 6.8\textdegree & 3.3 cm \vline\ 2.3\textdegree & \textbf{3.3 cm \vline\ 1.9\textdegree}                 & 3.3 cm \vline\ 2.0\textdegree                           \\
            25 cm/s  & 4.4 cm \vline\ 7.1\textdegree & 4.1 cm \vline\ 2.9\textdegree & \textbf{3.6 cm \vline\ 2.3\textdegree}                 & 4.3 cm \vline\ 2.6\textdegree                           \\
            \hline
            Training & 17h00 (20M)                   & 17h00 (20M)                   & 17h30 (20M)                                            & +82h00 (+1M)                                            \\
        \end{tabular}
    }
    \vspace{-0.3em}

    {\tiny DR -- domain randomization \quad | \quad PCG -- procedural content generation \quad | \quad PF -- particle fine-tuning}

    \vspace{-0.125em}
\end{table}

The baseline agent is trained in a static environment (Fig.~\ref{fig:sim_stacked}). It learns to solve the task efficiently in simulation but fails to generalize to the real world. The introduction of full domain randomization alone provides a substantial improvement in real-world performance. While the position error decreases moderately, the orientation error is reduced by more than half across all speeds. This demonstrates that randomizing physics, noise, and latency is a critical first step for creating a more robust policy.

While domain randomization is effective, the best overall generalization is yielded in combining it with PCG (Fig.~\ref{fig:sim_diverse}). Training agents on a unique terrain in each parallel instance achieves the lowest ATE at higher velocities. This result provides strong evidence for our central hypothesis. By preventing the agent from memorizing a single environment layout, procedural training forces it to learn the fundamental principles of dynamic navigation on varied terrain, leading to the most broadly effective policy.

The final experiment investigates whether performance can be enhanced by fine-tuning the policy with high-fidelity particle physics (Fig.~\ref{fig:sim_with_particles}). We take the median agent from the DR+PCG regime and continue its training for an additional 1M steps across eight parallel environments with particle simulation enabled. The results indicate that this step provides a slight improvement in ATE at the lowest velocity, but the benefit diminishes at higher speeds. This performance degradation at higher velocities can likely be attributed to a form of overfitting driven by reduced training diversity. The fine-tuning phase exposes the agent to a significantly narrower data distribution from only eight environments. This may cause the policy to lose some of the generalizability learned from the vast procedural diversity. Furthermore, while particle-based simulation offers higher fidelity, it may not perfectly capture all complex wheel-terrain interactions at high speeds. While this suggests that high-fidelity particle simulation can harden a policy against some unmodeled terramechanic effects, the significant training cost indicates that the broad generalization from PCG is a more critical and cost-effective strategy for this task.

\subsection{Stability through Action Smoothing}

While RL can produce highly performant policies, the resulting controllers often generate high-frequency, oscillating actions. These jerky commands can be effective for optimizing reward in simulation, but may lead to unstable behavior and cause excessive mechanical stress on physical hardware. This experiment investigates how different low-pass action filters can mitigate this issue. We compare the performance of the unfiltered DreamerV3 policy against versions augmented with a Moving Average filter (history of 5~steps), a third-order Savitzky-Golay filter (history of 9~steps), and a fourth-order Butterworth filter (cutoff frequency of 2.5~Hz). We evaluate both the ATE and the relative motion jerk, calculated as the time-averaged magnitude of the third derivative of the rover's position.

The results in \textsc{Table~\ref{tab:results_smoothing}} reveal a critical trade-off between tracking precision and motion stability. The unfiltered policy achieves the best tracking accuracy at higher speeds. However, this performance comes at the cost of a motion jerk three times higher than the filtered alternatives, making it unsuitable for long-term missions.

\begin{table}[t]
    \vspace{0.425em}
    \centering
    \caption{\textsc{Sim-to-Real Performance and Jerk of Action Smoothing Filters}}
    \label{tab:results_smoothing}
    \resizebox{\linewidth}{!}%
    {%
        \addtolength{\tabcolsep}{-0.3em}
        \begin{tabular}{@{}r|cccc@{}}
                    & \textbf{Unfiltered}                           & \textbf{Moving Average}                                 & \textbf{Savitzky-Golay}                   & \textbf{Butterworth}                      \\
            \hline
            5  cm/s & 2.3 cm \hspace{0.225mm}\vline\ 1.7\textdegree & \textbf{2.2 cm \hspace{-0.225mm}\vline\ 1.6\textdegree} & 2.6 cm \vline\ 1.7\textdegree             & 2.8 cm \vline\ 1.7\textdegree             \\
            15 cm/s & \textbf{3.3 cm \vline\ 1.9\textdegree}        & 3.7 cm \vline\ 2.4\textdegree                           & 5.0 cm \vline\ 2.3\textdegree             & 4.3 cm \vline\ 2.1\textdegree             \\
            25 cm/s & \textbf{3.6 cm \vline\ 2.3\textdegree}        & 4.2 cm \vline\ 2.1\textdegree                           & 64.9 cm \vline\ 16.4\textdegree           & 4.9 cm \vline\ 2.4\textdegree             \\
            \hline
            Jerk    & $100\hspace{-0.5mm}\pm\hspace{-0.5mm}81\%$    & $33\hspace{-0.5mm}\pm\hspace{-0.5mm}22\%$               & $30\hspace{-0.5mm}\pm\hspace{-0.5mm}20\%$ & $39\hspace{-0.5mm}\pm\hspace{-0.5mm}24\%$ \\
        \end{tabular}
    }
    \vspace{-1.0em}
\end{table}

As expected, all smoothing filters dramatically reduce jerk, leading to more predictable control. The Savitzky-Golay filter provides the greatest jerk reduction but introduces significant phase lag due to its wider history window, which causes its performance to degrade at 15~cm/s before failing catastrophically at 25~cm/s. With our setup, a simple Moving Average filter provides a satisfactory compromise. It improves tracking performance at low speed and accepts only a minor accuracy penalty at high speed, all while reducing mechanical jerk by 67\%. This experiment highlights that optimizing for raw performance alone is insufficient for hardware deployment. While other methods might be more effective, particularly those based on policy regularization, a well-tuned action filter represents a practical and computationally efficient method for achieving the stability required for safe and reliable long-term operation.

\subsection{Vision-Based Control}

\begin{figure}[b]
    \vspace{-0.5em}
    \centering
    \subcaptionbox{Simulated depth map.\label{fig:depth_sim}}{%
        \includegraphics[width=0.4935\linewidth]{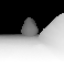}}
    \hfill
    \subcaptionbox{Real-world depth map.\label{fig:depth_real}}{%
        \includegraphics[width=0.4935\linewidth]{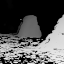}}
    \caption{Comparison of simulated and real-world depth views. The simulated map is clean, while the real-world image suffers from significant noise and signal dropout due to the properties of the basalt gravel present in the LunaLab.}
    \label{fig:depth}
    \vspace{-0.35em}
\end{figure}

Finally, we investigated the feasibility of end-to-end learning by training a policy with access to a $64\hspace{-0.3mm}\times\hspace{-0.3mm}64$~px depth map observation that is acquired from a RealSense D455 camera onboard the rover. While this policy could be transferred to the physical rover, its tracking accuracy was significantly degraded, with an ATE of 9.2~cm and 6.5\textdegree\ at 15~cm/s. We attribute this performance drop to a pronounced perceptual sim-to-real gap, visualized in Fig.~\ref{fig:depth}. Our standard simulation provides a clean depth map, but the physical basalt in LunaLab creates a noisy signal with substantial dropouts. This result highlights that even with a robust control policy, unmodeled sensor effects can be a primary failure point. Closing this perceptual gap through higher-fidelity sensor simulation is a critical challenge for future work.

\section{Conclusion}

This paper presented and validated a complete sim-to-real framework for rover navigation on challenging granular media. We demonstrated that a policy trained in procedurally diverse simulations can be deployed zero-shot to a physical rover, where it achieves precise and stable dynamic waypoint tracking in a lunar-analogue facility. Mastering this general capability is foundational for a wide range of applications, from following pre-planned science routes to assisting astronauts in the field. Our work provides a clear methodology for developing robust, learning-based controllers for such tasks in extraterrestrial traversal.

Our experiments yield several key findings for creating robust rover autonomy. We provided strong empirical evidence that training with procedural diversity is a critical factor for successful zero-shot transfer, as it forces the policy to generalize rather than overfit to simulation artifacts. We also found that model-based agents like DreamerV3 are particularly well-suited for this task due to their sample efficiency and learned world models. Furthermore, we demonstrated that simple action smoothing is a practical necessity for stable hardware deployment, and that a final fine-tuning with high-fidelity particle physics can offer marginal gains in low-speed precision at a significant training cost.

This work has two main limitations. First, our validation is confined to a single terrestrial analogue under Earth's gravity. Future work should validate these policies in more representative environments. Second, the framework relies on an external motion capture system for localization. Achieving full autonomy will require replacing this with onboard perception, a task whose difficulty was highlighted by the pronounced sim-to-real gap in our vision-based experiments.

This paper establishes a complete and validated workflow for creating robust learning-based navigation controllers. The demonstrated success of zero-shot transfer on challenging granular terrain marks a critical step towards creating autonomous systems that can reliably operate on the dusty surfaces of other worlds.

\balance

\vspace{1.0em}

\section*{Appendix: Hyperparameters}

\vspace{0.5em}

{
    \centering
    \resizebox{\linewidth}{!}%
    {%
        \begin{tabular}{@{}ll@{}}
            \hline
            \multicolumn{2}{l}{\textbf{PPO}}
            \vspace{-0.5em}                                                                                \\
            \quad Learning Rate (Actor \& Critic)       & \(0.0001 \xrightarrow[]{linear} 0.0\) (schedule) \\
            \quad Discount Factor ($\gamma$)            & \(0.997\)                                        \\
            \quad Rollout Buffer Size (per env)         & \(128\)                                          \\
            \quad Minibatch Size                        & \(1024\)                                         \\
            \quad PPO Epochs per Rollout                & \(16\)                                           \\
            \quad GAE Lambda ($\lambda$)                & \(0.95\)                                         \\
            \quad PPO Clip Range ($\epsilon$)           & \(0.2\)                                          \\
            \quad Entropy Coefficient                   & \(0.01\)                                         \\
            \quad Gradient Clipping Norm                & \(0.5\)                                          \\
            \quad Actor/Critic Network Size (MLP Units) & \([384, 384]\)                                   \\
            \hline
            \multicolumn{2}{l}{\textbf{PPO (LSTM)}}                                                        \\
            \quad LSTM Hidden Size                      & \(384\)                                          \\
            \hline
            \multicolumn{2}{l}{\textbf{TD3}}
            \vspace{-0.5em}                                                                                \\
            \quad Learning Rate (Actor \& Critic)       & \(0.003 \xrightarrow[]{linear} 0.0\) (schedule)  \\
            \quad Discount Factor ($\gamma$)            & \(0.997\)                                        \\
            \quad Replay Buffer Size                    & \(2,000,000\)                                    \\
            \quad Minibatch Size                        & \(512\)                                          \\
            \quad Updates per Environment Step          & \(4\)                                            \\
            \quad Exploration Noise                     & \(\mathcal{N}(0.0, 0.1)\)                        \\
            \quad Target Network Update Rate ($\tau$)   & \(0.005\)                                        \\
            \quad Actor/Critic Network Size (MLP Units) & \([384, 384]\)                                   \\
            \hline
            \multicolumn{2}{l}{\textbf{DreamerV3}}                                                         \\
            \quad Discount Factor ($\gamma$)            & \(0.997\)                                        \\
            \quad Replay Buffer Size                    & \(2,000,000\)                                    \\
            \quad Batch Size                            & \(16\)                                           \\
            \quad Sequence Length                       & \(64\)                                           \\
            \quad Updates per Environment Step          & \(32\)                                           \\
            \quad Model Size:                           &                                                  \\
            \quad \quad RSSM Hidden Size                & \(384\)                                          \\
            \quad \quad RSSM Deterministic Units        & \(3072\)                                         \\
            \quad \quad Discrete Latents per State      & \(24\)                                           \\
            \quad \quad MLP Units                       & \(384\)                                          \\
            \quad \quad CNN Depth                       & \(24\)                                           \\
            \hline
        \end{tabular}
    }
}


\begin{thebibliography}{10}
\providecommand{\url}[1]{#1}
\csname url@samestyle\endcsname
\providecommand{\newblock}{\relax}
\providecommand{\bibinfo}[2]{#2}
\providecommand{\BIBentrySTDinterwordspacing}{\spaceskip=0pt\relax}
\providecommand{\BIBentryALTinterwordstretchfactor}{4}
\providecommand{\BIBentryALTinterwordspacing}{\spaceskip=\fontdimen2\font plus
\BIBentryALTinterwordstretchfactor\fontdimen3\font minus
  \fontdimen4\font\relax}
\providecommand{\BIBforeignlanguage}[2]{{%
\expandafter\ifx\csname l@#1\endcsname\relax
\typeout{** WARNING: IEEEtran.bst: No hyphenation pattern has been}%
\typeout{** loaded for the language `#1'. Using the pattern for}%
\typeout{** the default language instead.}%
\else
\language=\csname l@#1\endcsname
\fi
#2}}
\providecommand{\BIBdecl}{\relax}
\BIBdecl

\bibitem{nasa2020artemis}
{National Aeronautics and Space Administration}, ``{Artemis Plan: NASA's Lunar
  Exploration Program Overview},'' 2020.

\bibitem{jpl2024enabling}
V.~Verma \emph{et~al.}, ``{Enabling Long \& Precise Drives for The Perseverance
  Mars Rover via Onboard Global Localization},'' in \emph{{IEEE Aerospace
  Conference}}, 2024, pp. 1--18.

\bibitem{zhang2019progress}
T.~Zhang \emph{et~al.}, ``{The Progress of Extraterrestrial Regolith-Sampling
  Robots},'' \emph{Nature Astronomy}, vol.~3, pp. 487--497, 2019.

\bibitem{sutton2018reinforcement}
R.~S. Sutton and A.~G. Barto, \emph{{Reinforcement Learning: An
  Introduction}}.\hskip 1em plus 0.5em minus 0.4em\relax A Bradford Book, 2018.

\bibitem{orsula2025space}
A.~Orsula, M.~Geist, M.~Olivares-Mendez, and C.~Martinez, ``{Space Robotics
  Bench: Robot Learning Beyond Earth},'' \emph{arXiv:2509.23328}, 2025.

\bibitem{ishigami2007terramechanics}
G.~Ishigami, A.~Miwa, K.~Nagatani, and K.~Yoshida, ``{Terramechanics-based
  model for steering maneuver of planetary exploration rovers on loose soil},''
  \emph{Journal of Field Robotics}, vol.~24, no.~3, pp. 233--250, 2007.

\bibitem{ishigami2007path}
G.~Ishigami, K.~Nagatani, and K.~Yoshida, ``{Path Planning for Planetary
  Exploration Rovers and Its Evaluation based on Wheel Slip Dynamics},'' in
  \emph{IEEE International Conference on Robotics and Automation}, 2007, pp.
  2361--2366.

\bibitem{ishigami2011path}
G.~Ishigami, K.~Nagatani, and K.~Yoshida, ``{Path Planning and Evaluation for
  Planetary Rovers Based on Dynamic Mobility Index},'' in \emph{IEEE/RSJ
  International Conference on Intelligent Robots and Systems}, 2011, pp.
  601--606.

\bibitem{tasora2015chrono}
A.~Tasora \emph{et~al.}, ``{Chrono: An open source multi-physics dynamics
  engine},'' in \emph{International Conference on High Performance Computing in
  Science and Engineering}, 2015, pp. 19--49.

\bibitem{xu2024reinforcement}
T.~Xu, C.~Pan, and X.~Xiao, ``{Reinforcement Learning for Wheeled Mobility on
  Vertically Challenging Terrain},'' in \emph{IEEE International Symposium on
  Safety Security Rescue Robotics}, 2024, pp. 125--130.

\bibitem{lu2024lunar}
S.~Lu, R.~Xu, Z.~Li, B.~Wang, and Z.~Zhao, ``{Lunar Rover Collaborated Path
  Planning with Artificial Potential Field-Based Heuristic on Deep
  Reinforcement Learning},'' \emph{Aerospace}, vol.~11, no.~4, 2024.

\bibitem{park2023deep}
B.-J. Park and H.-J. Chung, ``{Deep Reinforcement Learning-Based Failure-Safe
  Motion Planning for a 4-Wheeled 2-Steering Lunar Rover},'' \emph{Aerospace},
  vol.~10, no.~3, 2023.

\bibitem{mortensen2024twostage}
A.~B. Mortensen \emph{et~al.}, ``{Two-Stage Reinforcement Learning for
  Planetary Rover Navigation: Reducing the Reality Gap with Offline Noisy
  Data},'' in \emph{International Conference on Space Robotics}, 2024, pp.
  266--272.

\bibitem{cobbe2020leveraging}
K.~Cobbe, C.~Hesse, J.~Hilton, and J.~Schulman, ``{Leveraging procedural
  generation to benchmark reinforcement learning},'' in \emph{International
  Conference on Machine Learning}, 2020, pp. 2048--2056.

\bibitem{koutras2021marsexplorer}
D.~I. Koutras, A.~C. Kapoutsis, A.~A. Amanatiadis, and E.~B. Kosmatopoulos,
  ``{MarsExplorer: Exploration of Unknown Terrains via Deep Reinforcement
  Learning and Procedurally Generated Environments},'' \emph{Electronics},
  vol.~10, no.~22, 2021.

\bibitem{hughes2014verification}
S.~P. Hughes, R.~H. Qureshi, S.~D. Cooley, and J.~J. Parker, ``{Verification
  and Validation of the General Mission Analysis Tool (GMAT)},'' in
  \emph{AIAA/AAS Astrodynamics Specialist Conference}, 2014.

\bibitem{kenneally2020basilisk}
P.~W. Kenneally, S.~Piggott, and H.~Schaub, ``{Basilisk: A flexible, scalable
  and modular astrodynamics simulation framework},'' \emph{Journal of Aerospace
  Information Systems}, vol.~17, no.~9, pp. 496--507, 2020.

\bibitem{mortensen2024rlroverlab}
A.~B. Mortensen and S.~Bøgh, ``{RLRoverLAB: An Advanced Reinforcement Learning
  Suite for Planetary Rover Simulation and Training},'' in \emph{International
  Conference on Space Robotics}, 2024, pp. 273--277.

\bibitem{el2024drift}
M.~El-Hariry, A.~Richard, V.~Muralidharan, M.~Geist, and M.~Olivares-Mendez,
  ``{DRIFT: Deep Reinforcement Learning for Intelligent Floating Platforms
  Trajectories},'' in \emph{IEEE/RSJ International Conference on Intelligent
  Robots and Systems}, 2024, pp. 14\,034--14\,041.

\bibitem{wang2022collision}
S.~Wang, Y.~Cao, X.~Zheng, and T.~Zhang, ``{Collision-Free Trajectory Planning
  for a 6-DoF Free-Floating Space Robot via Hierarchical Decoupling
  Optimization},'' \emph{IEEE Robotics and Automation Letters}, vol.~7, no.~2,
  pp. 4953--4960, 2022.

\bibitem{tobin2017domain}
J.~Tobin \emph{et~al.}, ``{Domain randomization for transferring deep neural
  networks from simulation to the real world},'' in \emph{IEEE/RSJ
  International Conference on Intelligent Robots and Systems}, 2017, pp.
  23--30.

\bibitem{bousmalis2018using}
K.~Bousmalis \emph{et~al.}, ``{Using Simulation and Domain Adaptation to
  Improve Efficiency of Deep Robotic Grasping},'' in \emph{IEEE International
  Conference on Robotics and Automation}, 2018, pp. 4243--4250.

\bibitem{mittal2023orbit}
M.~Mittal \emph{et~al.}, ``{Orbit: A Unified Simulation Framework for
  Interactive Robot Learning Environments},'' \emph{IEEE Robotics and
  Automation Letters}, vol.~8, no.~6, pp. 3740--3747, 2023.

\bibitem{macenski2022ros2}
S.~Macenski, T.~Foote, B.~Gerkey, C.~Lalancette, and W.~Woodall, ``{Robot
  Operating System 2: Design, architecture, and uses in the wild},''
  \emph{Science Robotics}, vol.~7, no.~66, 2022.

\bibitem{ludivig2020building}
P.~Ludivig, A.~Calzada-Diaz, M.~Olivares-Mendez, H.~Voos, and J.~Lamamy,
  ``{Building a Piece of the Moon: Construction of Two Indoor Lunar Analogue
  Environments},'' in \emph{International Astronautical Congress}, 2020.

\bibitem{towers2024gymnasium}
M.~Towers \emph{et~al.}, ``{Gymnasium: A Standard Interface for Reinforcement
  Learning Environments},'' \emph{arXiv:2407.17032}, 2024.

\bibitem{schulman2017proximal}
J.~Schulman, F.~Wolski, P.~Dhariwal, A.~Radford, and O.~Klimov, ``{Proximal
  Policy Optimization Algorithms},'' \emph{arXiv:1707.06347}, 2017.

\bibitem{hochreiter1997long}
S.~Hochreiter and J.~Schmidhuber, ``{Long Short-Term Memory},'' \emph{Neural
  Computation}, vol.~9, no.~8, 1997.

\bibitem{fujimoto2018addressing}
S.~Fujimoto, H.~Hoof, and D.~Meger, ``{Addressing function approximation error
  in actor-critic methods},'' in \emph{International Conference on Machine
  Learning}, 2018, pp. 1587--1596.

\bibitem{hafner2025mastering}
D.~Hafner, J.~Pasukonis, J.~Ba, and T.~Lillicrap, ``{Mastering diverse control
  tasks through world models},'' \emph{Nature}, vol. 640, pp. 647--665, 2025.

\end{thebibliography}
\end{document}